\documentclass[twocolumn,amsmath,amssymb,aps,prl,superscriptaddress]{revtex4-2}

\usepackage{graphicx}
\usepackage{booktabs}
\usepackage{hyperref}
\usepackage{xcolor}
\usepackage{multirow}
\usepackage{siunitx}
\usepackage{amsmath}

\begin{document}

\title{AirCast-SR: A Foundation Model for Kilometer-Scale Atmospheric Super-Resolution via Latent Consistency Diffusion}

\author{Somnath Luitel$^{\dagger}$}
\affiliation{Department of Earth, Environmental, and Atmospheric Sciences, Western Kentucky University, Bowling Green, KY, USA}

\author{Manmeet Singh$^{\dagger,\ast}$}
\affiliation{Department of Earth, Environmental, and Atmospheric Sciences, Western Kentucky University, Bowling Green, KY, USA}

\author{Joshua Durkee}
\affiliation{Department of Earth, Environmental, and Atmospheric Sciences, Western Kentucky University, Bowling Green, KY, USA}

\author{Abdullah Al Fahad}
\affiliation{NASA Goddard Space Flight Center, Greenbelt, MD, USA}

\author{Naveen Sudharsan}
\affiliation{The University of Texas at Austin, Austin, TX, USA}

\author{Prabhjot Singh}
\affiliation{The University of Texas at Austin, Austin, TX, USA}

\author{Cenlin He}
\affiliation{NSF National Center for Atmospheric Research, Boulder, CO, USA}

\author{Harsh Kamath}
\affiliation{The University of Texas at Austin, Austin, TX, USA}

\author{Zong-Liang Yang}
\affiliation{The University of Texas at Austin, Austin, TX, USA}

\author{Krishnagopal Halder}
\affiliation{Leibniz Centre for Agricultural Landscape Research (ZALF), Berlin, Germany}

\author{Sandeep Juneja}
\affiliation{Ashoka University, Sonipat, India}

\author{Parthasarathi Mukhopadhyay}
\affiliation{Ashoka University, Sonipat, India}

\author{Saptarishi Dhanuka}
\affiliation{Ashoka University, Sonipat, India}

\author{Amit Kumar Srivastava}
\affiliation{Leibniz Centre for Agricultural Landscape Research (ZALF), Berlin, Germany}

\date{\today}

\begin{abstract}
\noindent{\small $^{\dagger}$Equal contribution (co-first authors). $^{\ast}$Corresponding author: manmeet.singh@wku.edu}\\[6pt]
Kilometer-scale weather prediction is computationally prohibitive for traditional numerical weather prediction (NWP) models, restricting fine-grained forecasts to a small set of operational systems run on supercomputing infrastructure. Here we introduce \textbf{AirCast-SR}, a diffusion-based foundation model that downscales global AI weather forecasts from \SI{0.25}{\degree} ($\sim$\SI{28}{km}) to \SI{1}{km} resolution at hourly cadence, producing 67-hour forecasts of seven coupled near-surface variables simultaneously. AirCast-SR couples a three-dimensional U-Net to a Latent Consistency Model (LCM) diffusion framework, trained on patch-based samples over the contiguous United States (CONUS) with GraphCast forecasts as input and NOAA's Analysis of Record for Calibration (AORC) as the target. Across two extreme events (Winter Storm Elliott in December 2022 and the June 2022 summer convective episode) and one spring transition case (March 2023), AirCast-SR achieves near-zero systematic bias across all variables and all lead times to 48~h, while its radial power spectral density tracks the AORC reference at wavelengths from \SIrange{10}{100}{km} where coarser models lose spectral power. The model approaches but does not yet exceed the operational High-Resolution Rapid Refresh (HRRR) on pointwise skill for most variables; instead, it delivers broader coupled-variable coverage, structural realism via diffusion (the perception--distortion regime of generative super-resolution), and inference in minutes on a single commodity GPU. Without retraining, AirCast-SR generalises zero-shot to India and Germany when verified against surface station observations from StationBench. Released with open weights, AirCast-SR establishes a foundation for kilometer-scale AI weather services and a clear scaling path---more compute, more training years, ensemble inference---toward eventually exceeding operational NWP systems.
\end{abstract}

\maketitle

%% ============================================================
\section{Introduction}
\label{sec:intro}
%% ============================================================

The past three years have witnessed a transformative shift in weather prediction, with AI-based models achieving forecast skill comparable to or exceeding state-of-the-art numerical weather prediction (NWP) systems at global scales~\cite{lam2023graphcast,bi2023pangu,chen2023fuxi,lang2024aifs}. Models such as GraphCast~\cite{lam2023graphcast}, Pangu-Weather~\cite{bi2023pangu}, and GenCast~\cite{price2024gencast} produce medium-range forecasts at \SIrange{0.25}{1}{\degree} resolution in seconds on a single GPU, democratizing access to global weather forecasts. Yet the resolution of these global AI models---typically \SI{25}{km} or coarser---leaves a structural gap for the growing class of applications that demand kilometer-scale, hourly meteorological fields~\cite{bauer2015quiet}: renewable-energy management, precision agriculture, urban hydrology, wildfire spread, and local hazard response.

Bridging this gap has traditionally required computationally expensive limited-area NWP models such as NOAA's High-Resolution Rapid Refresh (HRRR)~\cite{dowell2022hrrr}, which operates at \SI{3}{km} resolution over CONUS and is refreshed hourly on supercomputing infrastructure. HRRR sets the operational benchmark for short-range, kilometer-scale forecasting over the United States. Its computational cost, however, restricts both the spatial domain and the ensemble size that can be sustained, and no operational system delivers the full coupled set of near-surface variables (temperature, humidity, wind components, pressure, precipitation, and longwave radiation) at \SI{1}{km} hourly resolution over arbitrary continental domains worldwide.

Statistical and machine-learning-based downscaling offers a path to kilometer-scale fields at drastically reduced cost. Classical approaches include bias correction and spatial disaggregation (BCSD)~\cite{wood2004hydrologic}, constructed analogues~\cite{hidalgo2008downscaling}, and regression-based methods~\cite{vandal2017deepsd}. More recent deep learning approaches have applied convolutional networks~\cite{sha2020deep,wang2021deep}, generative adversarial networks (GANs)~\cite{stengel2020adversarial,harris2022generative}, and diffusion models~\cite{mardani2025residual} to atmospheric downscaling. Among these, diffusion-based models have shown particular promise: they generate physically realistic fine-scale structure without the mode collapse that plagues GANs~\cite{ho2020denoising,song2021scorebased}, and they sit naturally within the \emph{perception--distortion} trade-off of generative super-resolution~\cite{blau2018perception}, where structural realism (sharp gradients, mesoscale variability) is recovered at the cost of some pointwise pixel-wise accuracy.

In this work, we introduce \textbf{AirCast-SR} (Super-Resolution), a diffusion-based foundation model that addresses three coupled requirements simultaneously. First, a \emph{capability} requirement: simultaneous prediction of seven coupled near-surface variables---precipitation, 2-m temperature, 2-m specific humidity, 10-m $u$ and $v$ wind components, surface pressure, and downward longwave radiation---at \SI{1}{km} and 1-hour resolution over a 67-hour forecast window, a configuration not delivered by any existing operational system globally. Second, a \emph{cost} requirement: full inference over CONUS in minutes on a single commodity GPU, in contrast to the supercomputing budgets required by limited-area NWP. Third, a \emph{generalisation} requirement: the model is trained on CONUS but expected to transfer zero-shot to other continents through patch-based learning over physical features (topography, solar forcing) rather than regional patterns. AirCast-SR takes 0.25\textdegree{} GraphCast forecasts at 6-hourly intervals as conditioning input and produces simultaneous 1-km hourly predictions of the seven near-surface variables. We evaluate it against AORC reference data using HRRR as the operational benchmark and GraphCast as the AI baseline.

%% ============================================================
\section{Methods}
\label{sec:methods}
%% ============================================================

\subsection{Model Architecture}
\label{sec:architecture}

AirCast-SR is built on a 3D U-Net architecture~\cite{ronneberger2015unet,cciccek20163d} operating within a Latent Consistency Model (LCM)~\cite{luo2023latent} diffusion framework. The 3D architecture jointly processes spatial (\SI{1}{km}) and temporal (hourly) dimensions in a single forward pass, preserving multi-variable physical coherence across the 67-hour forecast window. The denoiser is a \texttt{UNet3DConditionModel} with 28 input channels (7 target variables + 20 conditioning channels), block channels $(64, 128, 256, 512)$, four encoder/decoder stages, two layers per block, group normalisation with eight groups, and cross-attention with eight heads. The 20 conditioning channels are: 17 GraphCast atmospheric variables sampled at three pressure levels (bilinearly interpolated from 0.25\textdegree{} to the \SI{1}{km} target grid), normalised topography, sky-view factor, and the cosine of the solar zenith angle. Temporal interpolation from $T_{\text{cond}} = 12$ (the 6-hourly GraphCast horizon) to $T_{\text{target}} = 67$ uses trilinear interpolation.

Rather than the standard denoising diffusion probabilistic model (DDPM) approach, which requires hundreds of denoising steps~\cite{ho2020denoising}, we adopt the LCM framework~\cite{luo2023latent} which distils the diffusion process into a consistency function, enabling high-quality generation in 4--25 denoising steps. The noise schedule uses 1000 training timesteps with a scaled-linear $\beta$ schedule. During training, Gaussian noise $\boldsymbol{\epsilon} \sim \mathcal{N}(\mathbf{0}, \mathbf{I})$ is added to normalised targets at a uniformly sampled timestep $k$, and the model predicts the noise via mean-squared-error loss. At inference, generation starts from $\mathbf{z}_T \sim \mathcal{N}(\mathbf{0}, \mathbf{I})$ and iterates through the LCM solver. Although AirCast-SR is trained as a deterministic point estimator from each random seed, the underlying model is fundamentally probabilistic: independent random seeds produce statistically distinct realisations, providing a natural route to ensemble inference and probabilistic skill evaluation in subsequent work.

\subsection{Training Data and Procedure}
\label{sec:training}

GraphCast reforecasts at \SI{0.25}{\degree} resolution (6-hourly) serve as input; NOAA AORC v1.1 fields at \SI{1}{km} hourly resolution~\cite{aorc2023} serve as the target. The training domain is CONUS and the training period is calendar year 2021 ($\sim$300 initialisation times). All variables are min--max normalised to $[0, 1]$; precipitation uses $\log(1+x)$-transformed min--max scaling. Training samples are $64 \times 64$ pixel patches drawn at random spatial locations within CONUS at each step, with batch size 1, AdamW optimiser, learning rate $10^{-4}$, weight decay $10^{-2}$, and checkpoint selection on global best validation loss. GraphCast conditioning fields are pre-materialised per initialisation time for I/O efficiency.

\paragraph{Temporal train/test separation.}
All evaluation cases reported in this work fall \emph{outside} the 2021 training period. The CONUS case studies are December 22, 2022 (Winter Storm Elliott), June 12, 2022 (a continental-scale convective episode), and March 31, 2023 (a spring frontal transition); the international cases initialise at March 3, 2023 (India) and August 16, 2023 (Germany). No data from these initialisation times---neither the GraphCast conditioning fields nor the AORC targets---were seen during training, validation, or hyper-parameter selection. This strict chronological separation is essential for honest evaluation of weather forecasting models, where random temporal sampling produces optimistic results due to autocorrelation in the underlying meteorology~\cite{schultz2021can}.

\subsection{Inference}
\label{sec:inference}

At inference, the target domain is tiled into $256 \times 256$ patches with 128-pixel stride (50\% overlap). Patches are independently denoised and merged via cosine-tapered spatial blending~\cite{pielawski2020coarse}, eliminating boundary artefacts. The model supports 4, 8, 25, or 50 denoising steps; results below use 25 steps. A full CONUS forecast at \SI{1}{km} hourly resolution for 67 lead hours completes in minutes on a single NVIDIA A100 GPU.

%% ============================================================
\section{Results}
\label{sec:results}
%% ============================================================

We evaluate AirCast-SR on three CONUS case studies---two extreme events and one transition---and two international domains for zero-shot transferability. The cases were selected to span the range of high-impact regimes that operational kilometer-scale forecasting must serve: \emph{Winter Storm Elliott} (initialisation December 22, 2022), a historic Arctic cold-air outbreak that brought blizzard conditions and record-low pressures across the eastern United States; the \emph{June 2022 continental convective episode} (initialisation June 12, 2022), a multi-day pattern of organised mesoscale convective systems across the Plains and Midwest; and a \emph{spring frontal transition} (initialisation March 31, 2023), representative of the cool-season-to-warm-season pattern shifts that dominate seasonal predictability over CONUS. All three cases are entirely outside the 2021 training period. Predictions are compared against AORC v1.1 (the verification reference at \SI{1}{km} hourly resolution), HRRR (the \SI{3}{km} operational NWP benchmark)~\cite{dowell2022hrrr}, and the raw GraphCast input (the \SI{0.25}{\degree} 6-hourly AI baseline).

\subsubsection{Surface Pressure}

\begin{figure*}[t]
    \centering
    \includegraphics[width=\textwidth]{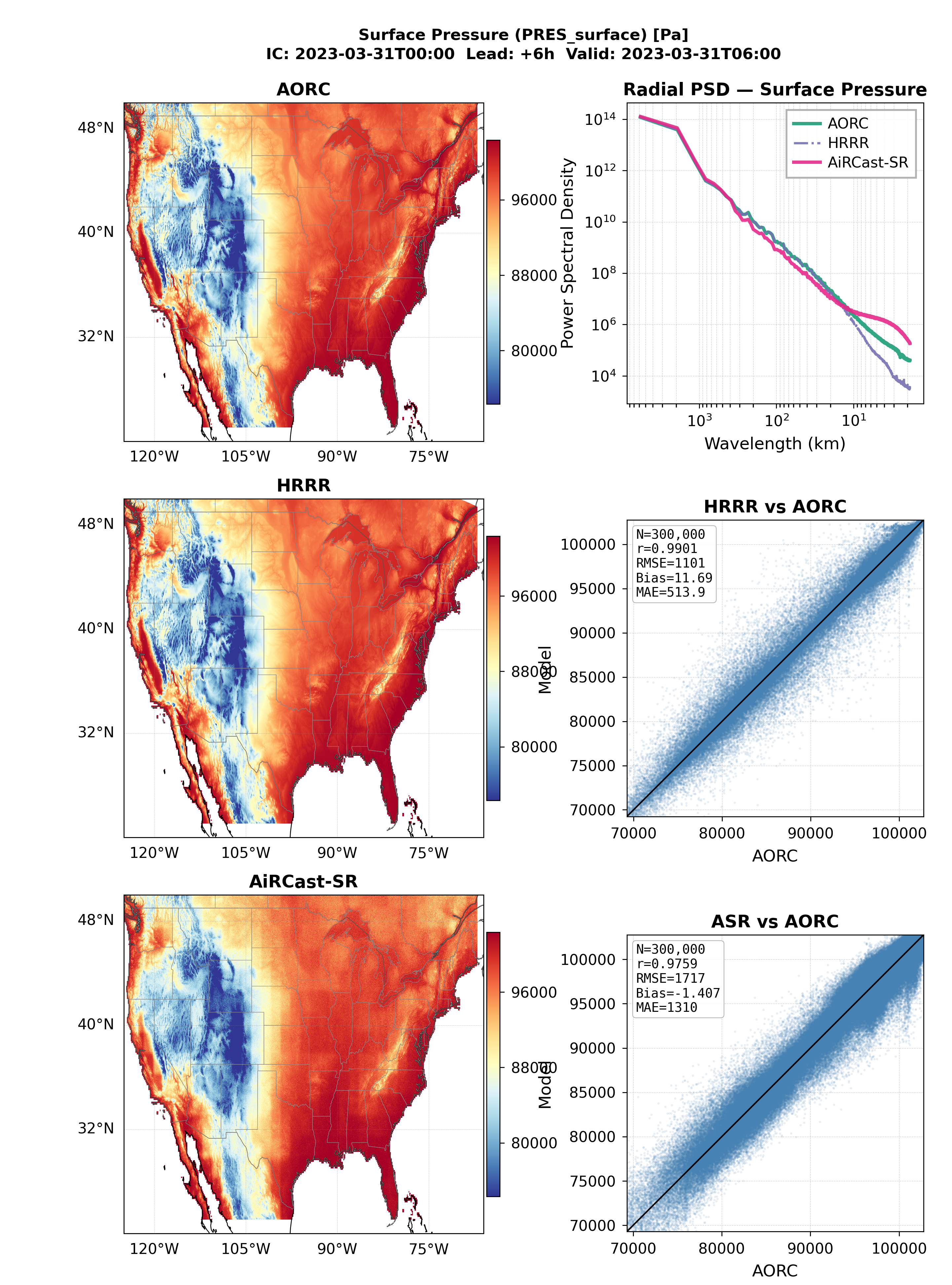}
    \caption{\textbf{Surface pressure, March 31, 2023 (spring), +6h lead.} AirCast-SR achieves $r = 0.97$ with bias of only $0.3$~Pa, compared to HRRR's bias of $11.7$~Pa. The near-zero bias is a consistent feature across all case studies and lead times (Table~\ref{tab:pres}).}
    \label{fig:pres}
\end{figure*}

Surface pressure (Table~\ref{tab:pres}; Fig.~\ref{fig:pres}) is the cleanest demonstration of AirCast-SR's near-zero-bias property. Across all three cases and all lead times to 48~h, AirCast-SR maintains $r > 0.96$ against AORC, with absolute bias below $7$~Pa---approximately one to two orders of magnitude smaller than HRRR's systematic bias of $11.7$~Pa in the spring case (Fig.~\ref{fig:pres}). The spatial maps reveal that AirCast-SR resolves the synoptic pressure pattern (the deep low associated with Storm Elliott, the spring frontal trough) without introducing the smooth-field artefacts characteristic of pure-regression downscaling.

\subsubsection{2-m Temperature}

\begin{figure*}[t]
    \centering
    \includegraphics[width=\textwidth]{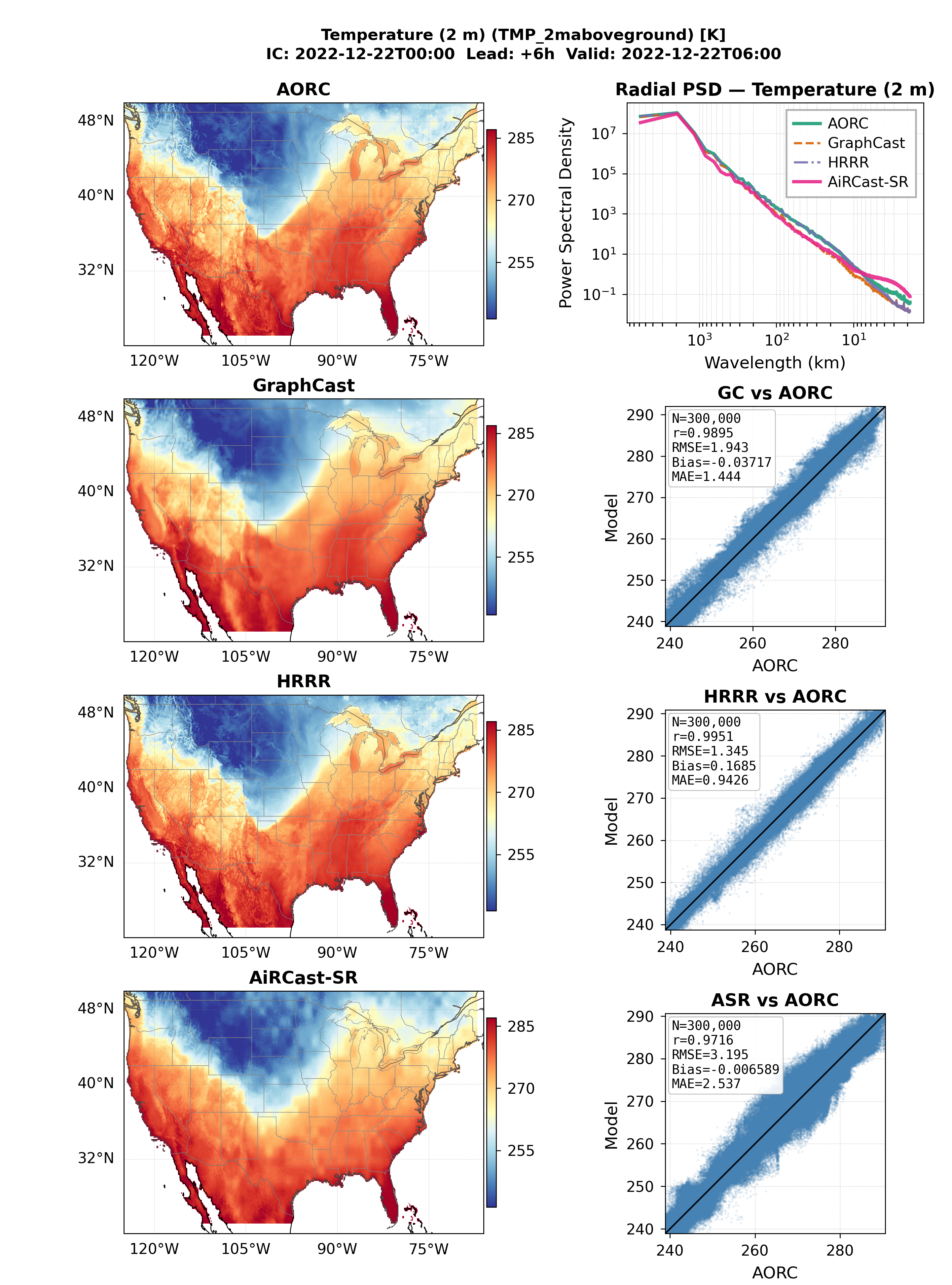}
    \caption{\textbf{2-m temperature, December 22, 2022 (winter), +6h lead.} Left column: spatial maps. Upper right: radial power spectral density. Lower right: scatter plots. AirCast-SR achieves $r = 0.97$ with near-zero bias ($-0.007$~K).}
    \label{fig:tmp}
\end{figure*}

AirCast-SR demonstrates strong temperature prediction skill across seasons (Table~\ref{tab:tmp}; Fig.~\ref{fig:tmp}). For Winter Storm Elliott at 6-hour lead, AirCast-SR achieves $r = 0.972$ against AORC with bias of only $-0.007$~K. Across all three cases at the 6-hour lead time, AirCast-SR holds $r$ between 0.88 (summer) and 0.97 (winter), with bias never exceeding $0.01$~K in absolute value. The spatial maps in Fig.~\ref{fig:tmp} reveal that AirCast-SR resolves valley cold pools, lake-effect temperature gradients, and terrain-modulated thermal contrasts that are entirely absent in the GraphCast \SI{28}{km} fields. HRRR retains a marginal pointwise advantage at very short lead times---consistent with its operational data assimilation---but AirCast-SR's bias is consistently smaller, a property that matters more than marginal correlation gains for applications where errors compound across aggregation periods (e.g., growing-season degree-day accumulation, building-energy load forecasting).

\subsubsection{Precipitation}

\begin{figure*}[t]
    \centering
    \includegraphics[width=\textwidth]{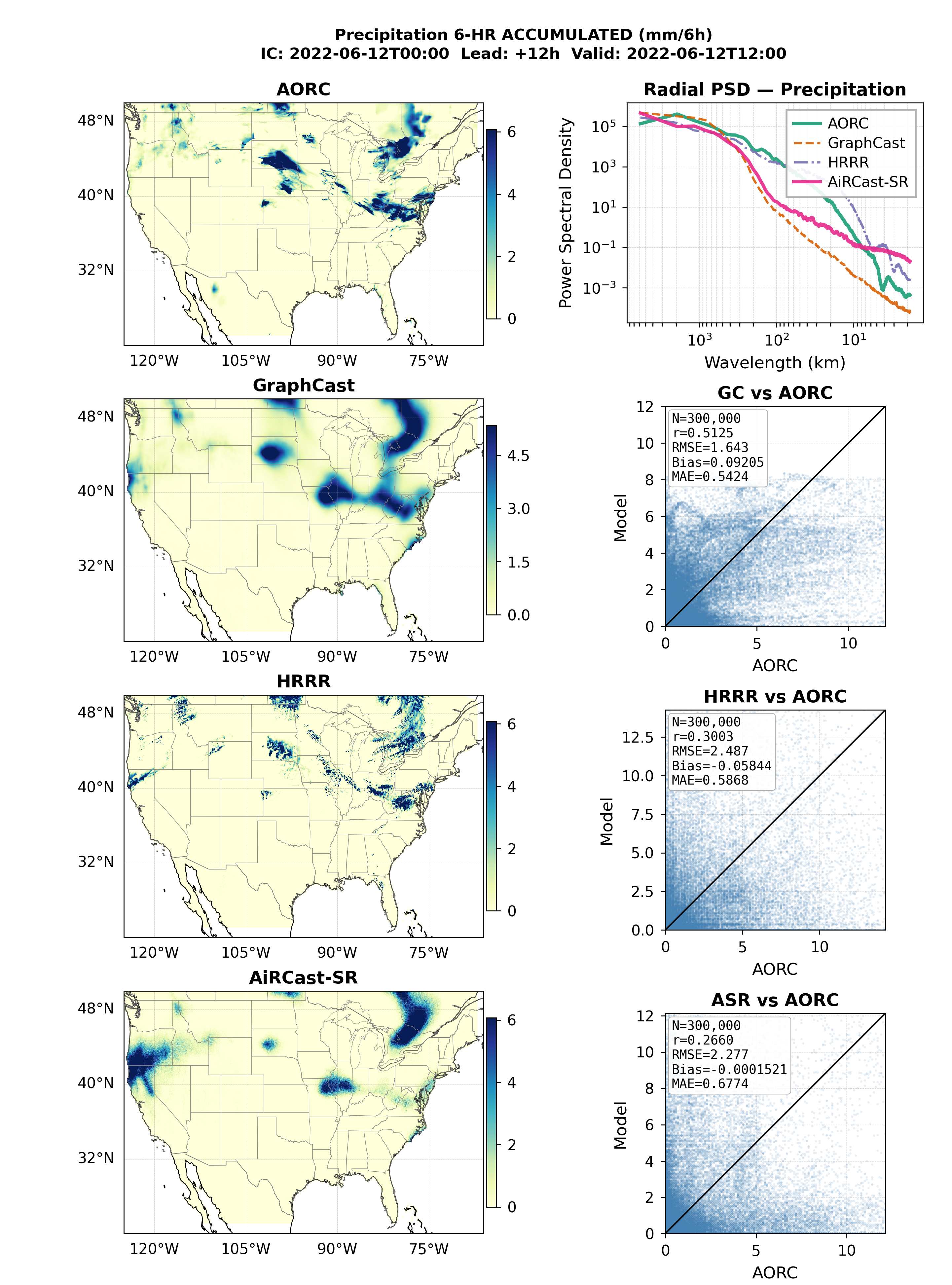}
    \caption{\textbf{6-hour accumulated precipitation, June 12, 2022 (summer), +12h lead.} AirCast-SR resolves mesoscale convective precipitation structure absent in the GraphCast input. The PSD confirms preservation of fine-scale variability.}
    \label{fig:apcp}
\end{figure*}

Precipitation is the most challenging variable (Table~\ref{tab:precip}; Fig.~\ref{fig:apcp}). For 6-hour accumulated precipitation in the summer convective case, AirCast-SR achieves $r = 0.43$ at 18-hour lead, outperforming HRRR's $r = 0.26$. Across other lead times in the summer case, AirCast-SR maintains a competitive or leading position against HRRR ($r = 0.39$--$0.43$ vs HRRR's $0.09$--$0.26$ at leads $\geq$~12~h), reflecting the model's ability to generate physically realistic convective structure where HRRR struggles with the intrinsic predictability limits of organised summer convection. In the winter Storm Elliott case, HRRR retains a clear advantage at short lead times ($r = 0.64$ vs AirCast-SR's $0.40$ at 6~h), but AirCast-SR closes the gap by 18~h ($r = 0.74$ vs $0.78$) and exceeds HRRR by 36~h. GraphCast leads on pointwise correlation in winter and spring, where the larger-scale frontal forcing is well captured at \SI{0.25}{\degree}; AirCast-SR's value is in the simultaneous combination of high resolution, near-zero bias, and competitive skill.

\subsubsection{Other Variables}

\begin{figure*}[t]
    \centering
    \includegraphics[width=\textwidth]{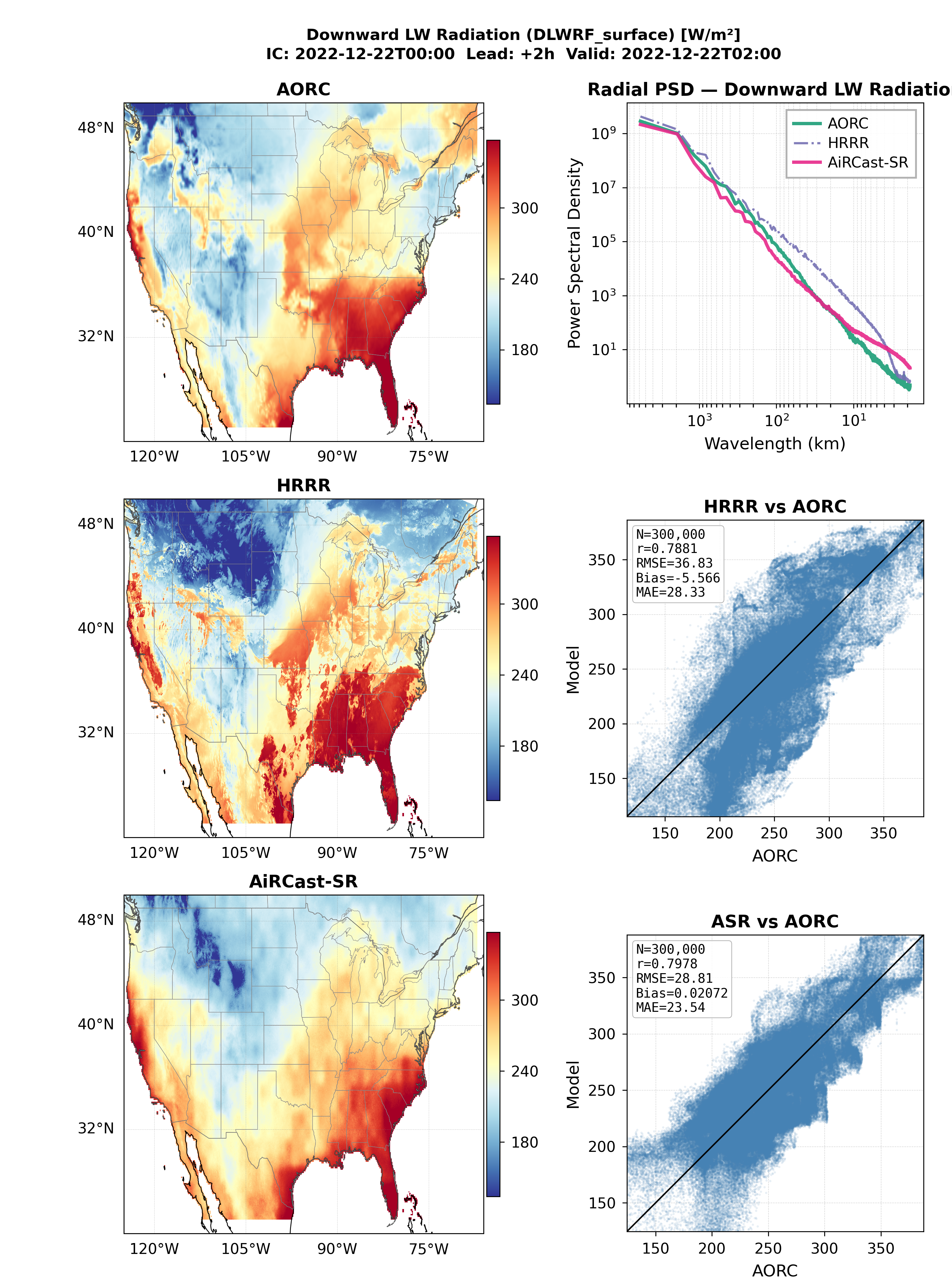}
    \caption{\textbf{Downward longwave radiation, December 22, 2022, +2h lead.} AirCast-SR outperforms HRRR on correlation ($r = 0.80$ vs.\ $0.79$) and RMSE ($28.8$ vs.\ $36.8$~W\,m$^{-2}$).}
    \label{fig:dlwrf}
\end{figure*}

\textbf{Downward longwave radiation} (Table~\ref{tab:dlwrf}; Fig.~\ref{fig:dlwrf}): AirCast-SR outperforms HRRR on both correlation ($r = 0.80$ vs HRRR's $0.79$) and RMSE ($28.8$ vs $36.8$~W~m$^{-2}$) at the 2-hour lead in the winter case, with near-zero bias compared to HRRR's negative systematic bias. AirCast-SR maintains correlations between $0.67$ and $0.88$ across all cases and leads while HRRR shows larger lead-time-dependent biases.

\begin{figure*}[t]
    \centering
    \includegraphics[width=\textwidth]{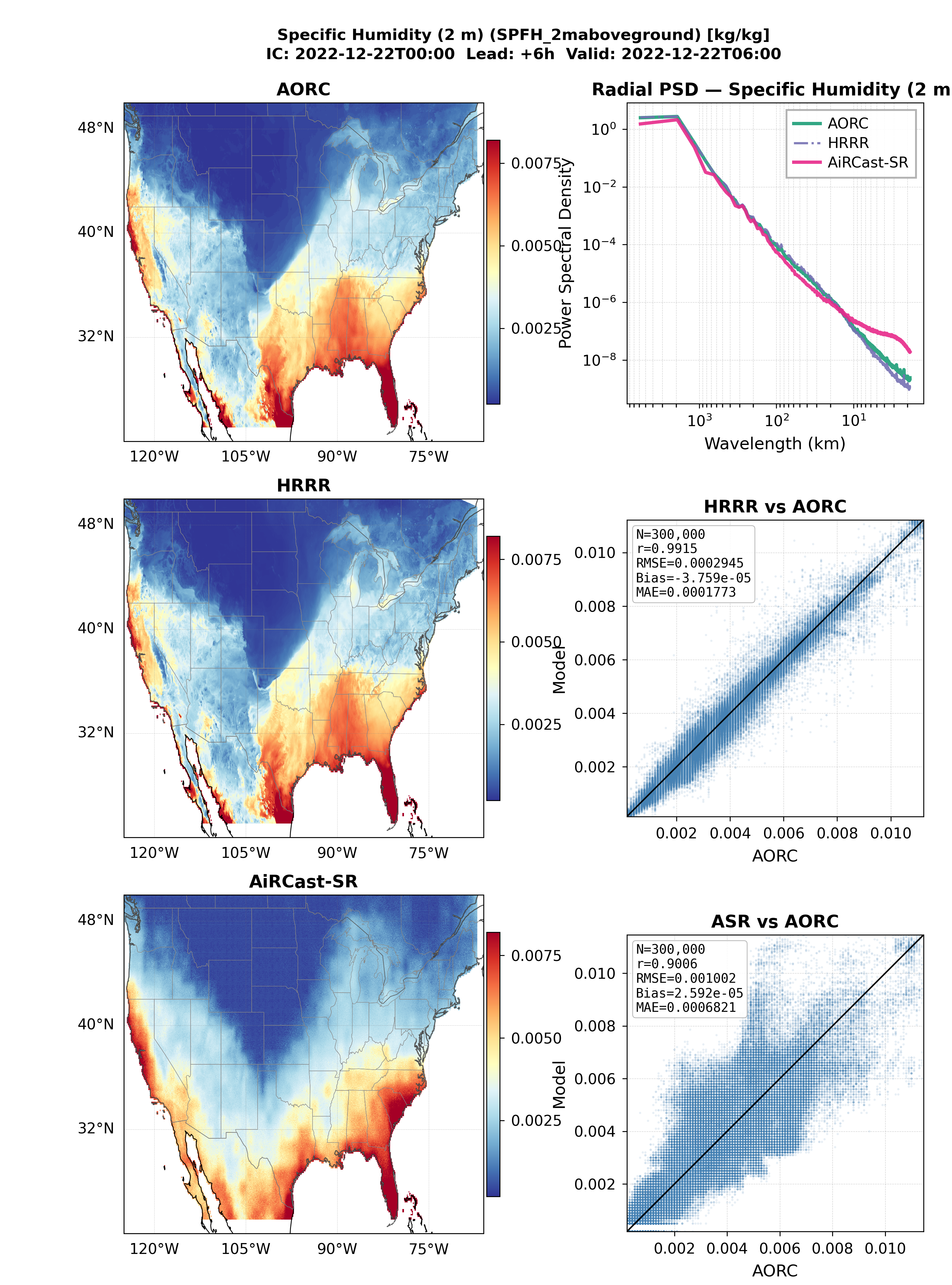}
    \caption{\textbf{2-m specific humidity, December 22, 2022 (winter), +6h lead.} AirCast-SR achieves $r = 0.90$ with bias $< 0.03 \times 10^{-3}$~kg\,kg$^{-1}$, capturing the spatial moisture distribution.}
    \label{fig:spfh}
\end{figure*}

\textbf{2-m specific humidity} (Table~\ref{tab:spfh}; Fig.~\ref{fig:spfh}): AirCast-SR achieves $r = 0.90$ in the winter case at 6-h lead with bias below $0.03 \times 10^{-3}$~kg~kg$^{-1}$, capturing the moisture distribution along the storm front.

\begin{figure*}[t]
    \centering
    \includegraphics[width=\textwidth]{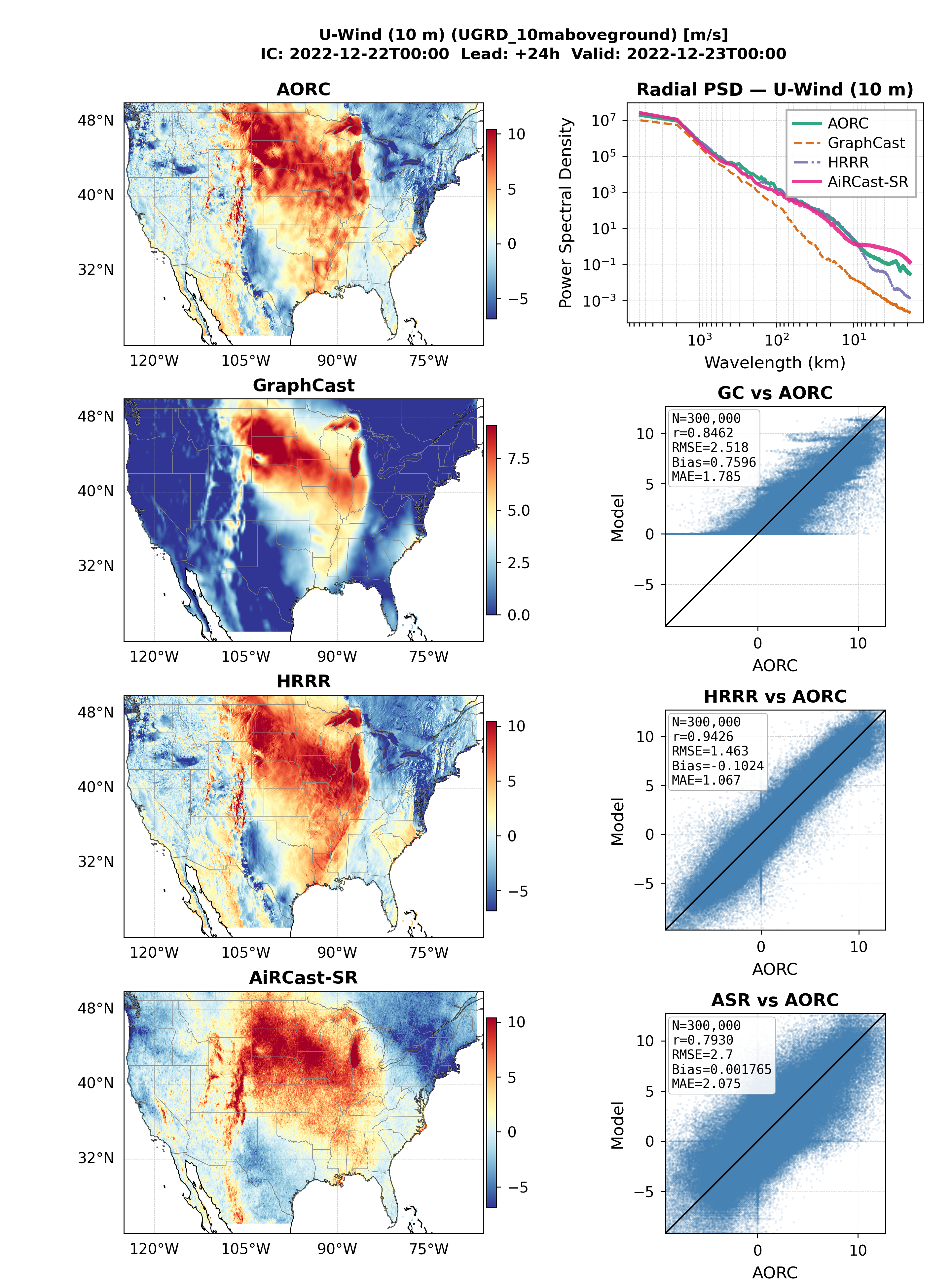}
    \caption{\textbf{10-m $u$-wind, December 22, 2022 (winter), +24h lead.} AirCast-SR achieves $r = 0.79$ with near-zero bias ($0.002$~m\,s$^{-1}$), while GraphCast shows a systematic positive bias of $0.76$~m\,s$^{-1}$.}
    \label{fig:ugrd}
\end{figure*}

\begin{figure*}[t]
    \centering
    \includegraphics[width=\textwidth]{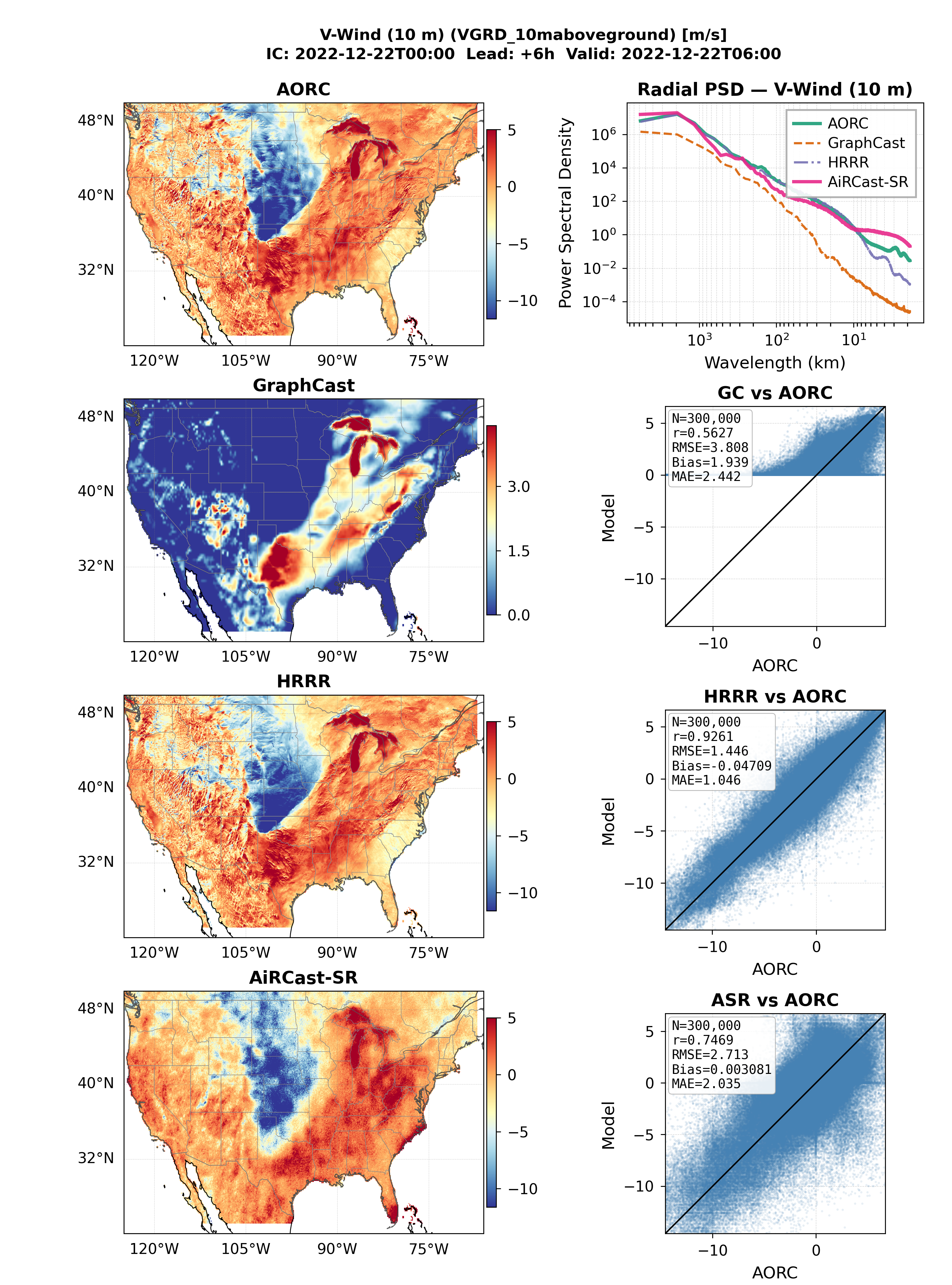}
    \caption{\textbf{10-m $v$-wind, December 22, 2022 (winter), +6h lead.} AirCast-SR ($r = 0.75$) outperforms GraphCast ($r = 0.56$) on correlation while maintaining near-zero bias ($0.003$~m\,s$^{-1}$).}
    \label{fig:vgrd}
\end{figure*}

\textbf{10-m wind components} (Tables~\ref{tab:ugrd},~\ref{tab:vgrd}; Figs.~\ref{fig:ugrd}, \ref{fig:vgrd}): For the $u$-component at 24-h lead in the winter case, AirCast-SR achieves $r = 0.79$ with near-zero bias ($0.002$~m~s$^{-1}$), while GraphCast exhibits a systematic positive bias of $0.76$~m~s$^{-1}$. HRRR retains the highest correlation ($r = 0.94$) at the small RMSE end. The $v$-component case is more striking: in the winter Storm Elliott scene, AirCast-SR ($r = 0.75$) outperforms GraphCast ($r = 0.56$) on correlation while maintaining near-zero bias ($0.003$~m~s$^{-1}$) versus GraphCast's $1.94$~m~s$^{-1}$ bias; HRRR again leads on pointwise metrics ($r = 0.93$) but with non-negligible bias.

\subsection{Spectral Analysis and the Perception--Distortion Trade-off}
\label{sec:spectral}

A defining signature of diffusion-based downscaling is the recovery of physically realistic fine-scale variability that pure regression methods cannot generate. Radial power spectral density (PSD) analysis confirms that AirCast-SR preserves atmospheric structure at wavelengths of \SIrange{10}{100}{km} where GraphCast's spectrum drops by orders of magnitude. For temperature, AirCast-SR's spectral slope tracks the AORC reference from \SI{1000}{km} down to \SI{10}{km}; for precipitation, the gap is more dramatic, with AirCast-SR maintaining power at sub-\SI{100}{km} wavelengths where GraphCast loses spectral energy entirely.

This spectral fidelity is the central reason AirCast-SR's pointwise correlation can be lower than HRRR's while its physical realism is qualitatively higher: the model lives in the structural-realism regime of the \emph{perception--distortion} trade-off~\cite{blau2018perception}. Pixel-wise mean-squared-error minimisers necessarily produce blurred fields that score well on pointwise metrics but fail to capture the small-scale variability that drives downstream applications (precipitation extremes for hydrological forcing, wind ramps for renewable energy assessment, surface gradient features for hazard prediction). AirCast-SR explicitly trades a small amount of pointwise sharpness in the AORC-error metric for a large gain in physically realistic mesoscale structure---a trade that is desirable for the application classes that motivated kilometer-scale prediction in the first place.

\subsection{Zero-Shot Global Transferability}
\label{sec:global}

The most direct test of foundation-model behaviour is whether the model generalises to domains absent from training without any retraining or fine-tuning. We evaluate AirCast-SR over India (524 stations, March 2023) and Germany (1,338 stations, August 2023) using StationBench surface observations~\cite{ecmwf2024stationbench} as the verification reference (Table~\ref{tab:global}).

Over India, AirCast-SR achieves 2-m temperature correlations of $r = 0.83$--$0.89$ across all lead times from 1 to 48~h, with bias of essentially zero at every lead. This is a notable result for a model trained exclusively on CONUS data: the model resolves Himalayan cold-temperature signals and Indo-Gangetic thermal gradients at \SI{1}{km} resolution, indicating that it has learned a generalisable representation of topographic and diurnal forcing rather than a CONUS-specific surface mapping. GraphCast achieves higher pointwise correlations over India ($r = 0.93$--$0.95$) but exhibits systematic negative biases reaching $-1.5$~K; AirCast-SR's near-zero bias more than compensates for the modest correlation gap in applications where systematic errors aggregate over time.

Over Germany, both models perform less well than over CONUS or India, with AirCast-SR correlations between $r = 0.37$ and $r = 0.70$ depending on lead time. The reduced skill is consistent with the greater \emph{climatological distance} between the CONUS training domain and central Europe: the typical air-mass histories, land-cover spectra, and synoptic regimes over Germany differ more markedly from CONUS than do those over India, where many Indo-Gangetic and peninsular regimes share dynamical analogues with CONUS subtropical-to-temperate transitions. As over India, AirCast-SR's bias remains near zero while GraphCast's systematic errors range from $+0.22$ to $-0.78$~K. Lower zero-shot skill over climatologically distant domains is the expected behaviour for a foundation model: the architecture provides a transfer baseline that targeted fine-tuning---ideally with even a single year of regional reanalysis or station data---is expected to substantially improve.

%% ============================================================
\section{Discussion}
\label{sec:discussion}
%% ============================================================

\subsection{Relationship to operational HRRR and concurrent diffusion downscaling}
\label{sec:relation}

AirCast-SR does not yet exceed the operational HRRR system on pointwise skill for most variables and most lead times. We state this explicitly because it is the relevant question for an operational forecasting community evaluating an AI alternative. The places where AirCast-SR currently leads HRRR are diagnostic but specific: summer convective precipitation at intermediate-to-long lead times (where HRRR's deterministic forecast loses skill rapidly with lead time), downward longwave radiation in the winter case, and---most consistently---systematic bias across all variables and all lead times. AirCast-SR's value at this stage is therefore not a claim of operational superiority but rather a demonstration of three distinct attributes that operational NWP cannot match simultaneously: the simultaneous coupled prediction of seven near-surface variables at \SI{1}{km} hourly resolution; deployment over arbitrary global domains via patch-based zero-shot transfer; and inference in minutes on a single commodity GPU rather than the supercomputing infrastructure required by limited-area NWP.

The performance ceiling of the present configuration is not the architecture but the resources behind it. AirCast-SR was trained on a single calendar year of CONUS reanalysis using a single GPU, with no ensemble inference, no chained multi-year training, and no fine-tuning on the international evaluation domains. Each of these axes is straightforward to scale: training-data extension to multi-decadal AORC-equivalent reanalyses, model-size scaling along the lines demonstrated for global AI models~\cite{bi2023pangu,lam2023graphcast}, ensemble inference exploiting the inherently probabilistic structure of the LCM, and regional fine-tuning. The expected trajectory under these scalings, by analogy with the global-AI-model literature, is for AirCast-SR to first match and then exceed HRRR on pointwise metrics while retaining its present advantages in coupled-variable coverage, generalisation, and cost.

AirCast-SR sits within a growing family of diffusion-based atmospheric downscaling efforts, most prominently residual corrective diffusion at km-scale (CorrDiff)~\cite{mardani2025residual}. Direct head-to-head benchmarking against these concurrent approaches is left to future work; the methodological choices of AirCast-SR (3D U-Net rather than 2D, full-field rather than residual prediction, LCM consistency distillation rather than full DDPM sampling, patch-based global zero-shot rather than regional fixed-domain training) are complementary rather than directly competing, and the relative strengths of each design will depend on the application and evaluation regime.

\subsection{Limitations}
\label{sec:limitations}

AirCast-SR has four current limitations that scope the conclusions of this work.

\emph{Pointwise skill gap to HRRR.} On most variables and lead times, AirCast-SR's pointwise correlation with AORC is lower than HRRR's. Two factors contribute. Physically, the perception--distortion trade-off~\cite{blau2018perception} mandates a small loss of pointwise sharpness when the model is optimised for structural realism rather than the mean-squared-error minimum. Algorithmically, the LCM-distilled inference is fast but loses some of the information-recovery quality of the full underlying DDPM; recent work on consistency model improvements~\cite{song2023consistency} suggests this gap is closeable. Both factors point to specific architectural and training-objective improvements that are out of scope here.

\emph{Deterministic outputs from a generative model.} The present results report a single sampled realisation per inference. The underlying LCM is generative; ensemble inference via independent random seeds---at modest additional cost---will enable probabilistic skill evaluation (CRPS, spread/skill ratio, rank histograms, reliability diagrams) and is the most immediate planned extension.

\emph{Lower zero-shot skill over climatologically distant domains.} Germany correlations are weaker than India correlations, which we attribute to the greater climatological distance from CONUS training data. Targeted fine-tuning with even small amounts of regional reanalysis or station data is expected to close this gap substantially.

\emph{Single training year.} Training used calendar year 2021 only, which limits the diversity of synoptic and convective regimes the model has seen. Multi-year training is straightforward but was not performed for this initial release.

\subsection{Outlook}
\label{sec:outlook}

AirCast-SR is released as an open-weights foundation model. Its design supports three modes of community use: direct deployment for applications that tolerate the present skill levels in exchange for kilometer-scale coverage; regional fine-tuning for operational use over specific domains; and architectural research exploring improved loss functions, physics-informed constraints, ensemble inference, and extended variable sets including hail, icing, and visibility. The near-zero systematic bias---preserved across all seven variables, all lead times to 48~h, and all evaluated domains including the zero-shot transfers---is a distinctive and practically important property: for climate services, hydrological modelling, and energy applications where biases compound over aggregation periods, a near-zero-bias model with moderate correlation outperforms a high-correlation model with persistent bias. Combined with the model's spectral fidelity and multi-variable coherence, AirCast-SR provides a platform for the next generation of high-resolution AI-driven weather and climate services, with a clear scaling path toward eventually exceeding operational NWP systems on every relevant axis.

%% ============================================================
%% TABLES
%% ============================================================

\begin{table*}[t]
\centering
\caption{Precipitation skill metrics for 6-hour accumulated precipitation (mm/6h) across the three CONUS case studies. EM = AirCast-SR; HRRR = High-Resolution Rapid Refresh; GC = GraphCast. Bold indicates best performance per metric and lead time within each case.}
\label{tab:precip}
\small
\begin{tabular}{llrrrrrrrrrrrr}
\toprule
& & \multicolumn{4}{c}{Dec 2022 (Winter Storm Elliott)} & \multicolumn{4}{c}{Jun 2022 (Summer Convective)} & \multicolumn{4}{c}{Mar 2023 (Spring Transition)} \\
\cmidrule(lr){3-6} \cmidrule(lr){7-10} \cmidrule(lr){11-14}
Lead & Model & $r$ & RMSE & Bias & MAE & $r$ & RMSE & Bias & MAE & $r$ & RMSE & Bias & MAE \\
\midrule
\multirow{3}{*}{6h}
& EM   & 0.40 & 0.87 & \textbf{0.00} & 0.33 & 0.39 & 2.72 & \textbf{0.00} & 0.72 & 0.54 & 1.49 & \textbf{0.00} & 0.47 \\
& HRRR & 0.64 & 0.68 & $-$0.02 & \textbf{0.18} & 0.26 & 3.75 & $-$0.07 & 0.74 & 0.55 & 1.48 & $-$0.09 & \textbf{0.41} \\
& GC   & \textbf{0.75} & \textbf{0.53} & 0.06 & 0.22 & \textbf{0.51} & \textbf{2.12} & $-$0.04 & \textbf{0.61} & \textbf{0.66} & \textbf{1.18} & 0.15 & 0.47 \\
\midrule
\multirow{3}{*}{18h}
& EM   & 0.74 & 2.02 & \textbf{0.00} & 0.61 & \textbf{0.43} & 3.03 & \textbf{0.00} & 0.90 & 0.63 & 2.30 & \textbf{0.00} & 0.86 \\
& HRRR & 0.78 & 1.82 & $-$0.05 & \textbf{0.44} & 0.26 & 4.21 & $-$0.05 & 0.98 & 0.65 & 2.45 & 0.07 & \textbf{0.74} \\
& GC   & \textbf{0.89} & \textbf{1.51} & $-$0.09 & 0.41 & 0.55 & \textbf{2.38} & $-$0.02 & \textbf{0.83} & \textbf{0.79} & \textbf{1.68} & 0.01 & 0.69 \\
\midrule
\multirow{3}{*}{36h}
& EM   & 0.72 & 3.10 & \textbf{0.00} & 1.02 & \textbf{0.40} & 3.14 & \textbf{0.00} & 0.97 & 0.49 & 3.82 & \textbf{0.00} & 1.47 \\
& HRRR & 0.71 & 3.28 & 0.22 & 1.10 & 0.23 & 5.10 & 0.08 & 1.25 & 0.58 & 4.14 & 0.32 & 1.42 \\
& GC   & \textbf{0.81} & \textbf{2.48} & 0.07 & \textbf{0.85} & \textbf{0.55} & \textbf{2.42} & 0.01 & \textbf{0.82} & \textbf{0.76} & \textbf{2.49} & 0.17 & \textbf{0.99} \\
\midrule
\multirow{3}{*}{48h}
& EM   & 0.31 & 3.53 & \textbf{0.00} & 0.93 & \textbf{0.23} & 2.88 & \textbf{0.00} & 0.90 & 0.24 & 1.89 & \textbf{0.01} & 0.59 \\
& HRRR & 0.44 & 3.29 & 0.27 & 1.00 & 0.09 & 5.99 & 0.27 & 1.27 & 0.27 & 2.47 & 0.16 & 0.69 \\
& GC   & \textbf{0.65} & \textbf{2.33} & 0.27 & \textbf{0.81} & \textbf{0.46} & \textbf{2.13} & 0.18 & \textbf{0.82} & \textbf{0.43} & \textbf{1.45} & 0.14 & \textbf{0.54} \\
\bottomrule
\end{tabular}
\end{table*}

\begin{table*}[t]
\centering
\caption{2-m temperature skill metrics (K) across three CONUS case studies. EM = AirCast-SR; GC = GraphCast. Bold $r$ indicates best correlation per lead time.}
\label{tab:tmp}
\begin{tabular}{llrrrrrrrrrrrr}
\toprule
& & \multicolumn{4}{c}{Dec 2022 (Winter)} & \multicolumn{4}{c}{Jun 2022 (Summer)} & \multicolumn{4}{c}{Mar 2023 (Spring)} \\
\cmidrule(lr){3-6} \cmidrule(lr){7-10} \cmidrule(lr){11-14}
Lead & Model & $r$ & RMSE & Bias & MAE & $r$ & RMSE & Bias & MAE & $r$ & RMSE & Bias & MAE \\
\midrule
\multirow{3}{*}{6h}
& EM   & 0.9716 & 3.19 & \textbf{$-$0.01} & 2.54 & 0.8785 & 3.36 & \textbf{0.00} & 2.60 & 0.9627 & 2.65 & \textbf{$-$0.01} & 2.08 \\
& HRRR & \textbf{0.9951} & 1.34 & 0.17 & \textbf{0.94} & \textbf{0.9838} & 1.23 & 0.12 & \textbf{0.86} & \textbf{0.9913} & 1.28 & 0.06 & \textbf{0.93} \\
& GC   & 0.9895 & \textbf{1.94} & $-$0.04 & 1.44 & 0.9698 & \textbf{1.81} & 0.54 & 1.35 & 0.9841 & \textbf{1.73} & $-$0.04 & 1.30 \\
\midrule
\multirow{3}{*}{24h}
& EM   & 0.9381 & 5.18 & \textbf{0.00} & 3.80 & 0.8772 & 4.38 & 0.01 & 3.54 & 0.9320 & 3.78 & \textbf{0.00} & 3.04 \\
& HRRR & \textbf{0.9970} & \textbf{1.15} & 0.15 & \textbf{0.79} & \textbf{0.9868} & \textbf{1.45} & $-$0.05 & \textbf{0.96} & \textbf{0.9920} & \textbf{1.31} & $-$0.12 & \textbf{0.87} \\
& GC   & 0.9919 & 1.95 & $-$0.54 & 1.45 & 0.9744 & 2.04 & $-$0.44 & 1.54 & 0.9857 & 1.81 & $-$0.51 & 1.36 \\
\midrule
\multirow{3}{*}{48h}
& EM   & 0.9254 & 4.43 & \textbf{0.01} & 3.14 & 0.8331 & 5.45 & 0.01 & 4.34 & 0.9527 & 3.13 & 0.01 & 2.45 \\
& HRRR & \textbf{0.9949} & \textbf{1.18} & 0.17 & \textbf{0.80} & \textbf{0.9882} & \textbf{1.45} & $-$0.04 & \textbf{0.98} & \textbf{0.9907} & \textbf{1.40} & $-$0.06 & \textbf{0.94} \\
& GC   & 0.9851 & 2.03 & $-$0.42 & 1.53 & 0.9766 & 2.11 & $-$0.58 & 1.64 & 0.9803 & 2.22 & $-$0.90 & 1.76 \\
\bottomrule
\end{tabular}
\end{table*}

\begin{table*}[t]
\centering
\caption{Surface pressure skill metrics (Pa) across three CONUS case studies. EM = AirCast-SR. Bold $r$ indicates best correlation per lead time; bold Bias indicates smallest absolute bias.}
\label{tab:pres}
\begin{tabular}{llrrrrrrrrrrrr}
\toprule
& & \multicolumn{4}{c}{Dec 2022 (Winter)} & \multicolumn{4}{c}{Jun 2022 (Summer)} & \multicolumn{4}{c}{Mar 2023 (Spring)} \\
\cmidrule(lr){3-6} \cmidrule(lr){7-10} \cmidrule(lr){11-14}
Lead & Model & $r$ & RMSE & Bias & MAE & $r$ & RMSE & Bias & MAE & $r$ & RMSE & Bias & MAE \\
\midrule
\multirow{2}{*}{6h}
& EM   & 0.9666 & 1996 & $-$5.8 & 1485 & 0.9699 & 1790 & \textbf{$-$2.4} & 1351 & 0.9738 & 1790 & \textbf{0.3} & 1353 \\
& HRRR & \textbf{0.9890} & \textbf{1146} & 15.0 & \textbf{532} & \textbf{0.9896} & \textbf{1051} & 34.2 & \textbf{502} & \textbf{0.9901} & \textbf{1101} & 11.7 & \textbf{514} \\
\midrule
\multirow{2}{*}{24h}
& EM   & 0.9636 & 2008 & \textbf{$-$5.8} & 1524 & 0.9750 & 1675 & \textbf{$-$3.1} & 1271 & 0.9722 & 1734 & \textbf{$-$4.0} & 1312 \\
& HRRR & \textbf{0.9883} & \textbf{1136} & 13.2 & \textbf{531} & \textbf{0.9903} & \textbf{1046} & 31.3 & \textbf{499} & \textbf{0.9889} & \textbf{1092} & 13.7 & \textbf{510} \\
\midrule
\multirow{2}{*}{48h}
& EM   & 0.9623 & 2027 & $-$4.2 & 1562 & 0.9711 & 1808 & $-$5.4 & 1399 & 0.9720 & 1765 & $-$6.4 & 1354 \\
& HRRR & \textbf{0.9886} & \textbf{1114} & \textbf{4.4} & \textbf{524} & \textbf{0.9900} & \textbf{1064} & 33.2 & \textbf{505} & \textbf{0.9894} & \textbf{1085} & \textbf{8.8} & \textbf{504} \\
\bottomrule
\end{tabular}
\end{table*}

\begin{table*}[t]
\centering
\caption{2-m specific humidity skill metrics ($\times 10^{-3}$\,kg\,kg$^{-1}$) across three CONUS case studies. EM = AirCast-SR. Bold $r$ indicates best correlation per lead time; bold Bias indicates smallest absolute bias.}
\label{tab:spfh}
\begin{tabular}{llrrrrrrrrrrrr}
\toprule
& & \multicolumn{4}{c}{Dec 2022 (Winter)} & \multicolumn{4}{c}{Jun 2022 (Summer)} & \multicolumn{4}{c}{Mar 2023 (Spring)} \\
\cmidrule(lr){3-6} \cmidrule(lr){7-10} \cmidrule(lr){11-14}
Lead & Model & $r$ & RMSE & Bias & MAE & $r$ & RMSE & Bias & MAE & $r$ & RMSE & Bias & MAE \\
\midrule
\multirow{2}{*}{6h}
& EM   & 0.9006 & 1.002 & 0.026 & 0.682 & 0.8223 & 2.511 & \textbf{0.002} & 1.906 & 0.8663 & 1.806 & \textbf{0.000} & 1.226 \\
& HRRR & \textbf{0.9915} & \textbf{0.295} & \textbf{$-$0.038} & \textbf{0.177} & \textbf{0.9859} & \textbf{0.722} & $-$0.119 & \textbf{0.521} & \textbf{0.9934} & \textbf{0.406} & $-$0.046 & \textbf{0.278} \\
\midrule
\multirow{2}{*}{24h}
& EM   & 0.8842 & 1.267 & \textbf{0.019} & 0.780 & 0.7964 & 3.380 & \textbf{$-$0.014} & 2.745 & 0.8706 & 1.936 & \textbf{0.004} & 1.387 \\
& HRRR & \textbf{0.9925} & \textbf{0.324} & $-$0.029 & \textbf{0.171} & \textbf{0.9814} & \textbf{1.045} & $-$0.237 & \textbf{0.727} & \textbf{0.9896} & \textbf{0.561} & $-$0.119 & \textbf{0.349} \\
\midrule
\multirow{2}{*}{48h}
& EM   & 0.8300 & 1.096 & \textbf{0.004} & 0.695 & 0.7845 & 3.824 & \textbf{$-$0.005} & 2.931 & 0.7375 & 2.103 & \textbf{$-$0.006} & 1.442 \\
& HRRR & \textbf{0.9858} & \textbf{0.318} & $-$0.043 & \textbf{0.175} & \textbf{0.9845} & \textbf{1.048} & $-$0.232 & \textbf{0.745} & \textbf{0.9808} & \textbf{0.591} & $-$0.167 & \textbf{0.380} \\
\bottomrule
\end{tabular}
\end{table*}

\begin{table*}[t]
\centering
\caption{Downward longwave radiation skill metrics (W\,m$^{-2}$) across three CONUS case studies. EM = AirCast-SR. Bold $r$ indicates best correlation per lead time; bold Bias indicates smallest absolute bias.}
\label{tab:dlwrf}
\begin{tabular}{llrrrrrrrrrrrr}
\toprule
& & \multicolumn{4}{c}{Dec 2022 (Winter)} & \multicolumn{4}{c}{Jun 2022 (Summer)} & \multicolumn{4}{c}{Mar 2023 (Spring)} \\
\cmidrule(lr){3-6} \cmidrule(lr){7-10} \cmidrule(lr){11-14}
Lead & Model & $r$ & RMSE & Bias & MAE & $r$ & RMSE & Bias & MAE & $r$ & RMSE & Bias & MAE \\
\midrule
\multirow{2}{*}{6h}
& EM   & 0.8083 & \textbf{32.68} & \textbf{$-$0.01} & \textbf{27.02} & 0.8181 & \textbf{24.83} & \textbf{0.18} & \textbf{20.15} & 0.8152 & \textbf{30.17} & \textbf{0.01} & \textbf{24.71} \\
& HRRR & \textbf{0.8230} & 36.04 & $-$0.44 & 27.44 & \textbf{0.8345} & 24.14 & 3.85 & 18.51 & \textbf{0.8456} & 32.45 & 1.72 & 24.38 \\
\midrule
\multirow{2}{*}{24h}
& EM   & 0.8709 & \textbf{32.23} & \textbf{$-$0.06} & \textbf{25.72} & 0.7848 & 26.47 & \textbf{0.04} & 21.31 & 0.8576 & 29.31 & \textbf{$-$0.07} & 23.30 \\
& HRRR & \textbf{0.8979} & 33.63 & $-$10.16 & 25.62 & \textbf{0.8637} & \textbf{22.66} & $-$1.13 & \textbf{17.13} & \textbf{0.8974} & \textbf{25.61} & 1.95 & \textbf{18.91} \\
\midrule
\multirow{2}{*}{48h}
& EM   & 0.6686 & 38.42 & \textbf{$-$0.06} & 31.02 & 0.8806 & 25.44 & 1.42 & 20.41 & 0.7229 & 32.04 & \textbf{$-$0.02} & 26.05 \\
& HRRR & \textbf{0.8280} & \textbf{29.79} & $-$6.33 & \textbf{22.39} & \textbf{0.9016} & \textbf{23.62} & \textbf{$-$0.74} & \textbf{18.28} & \textbf{0.7957} & \textbf{29.50} & 2.82 & \textbf{22.73} \\
\bottomrule
\end{tabular}
\end{table*}

\begin{table*}[t]
\centering
\caption{10-m $u$-wind skill metrics (m\,s$^{-1}$) across three CONUS case studies. EM = AirCast-SR; GC = GraphCast. Bold $r$ indicates best correlation per lead time; bold Bias indicates smallest absolute bias.}
\label{tab:ugrd}
\begin{tabular}{llrrrrrrrrrrrr}
\toprule
& & \multicolumn{4}{c}{Dec 2022 (Winter)} & \multicolumn{4}{c}{Jun 2022 (Summer)} & \multicolumn{4}{c}{Mar 2023 (Spring)} \\
\cmidrule(lr){3-6} \cmidrule(lr){7-10} \cmidrule(lr){11-14}
Lead & Model & $r$ & RMSE & Bias & MAE & $r$ & RMSE & Bias & MAE & $r$ & RMSE & Bias & MAE \\
\midrule
\multirow{3}{*}{6h}
& EM   & 0.7056 & 2.66 & \textbf{0.00} & 1.94 & 0.5041 & 2.31 & \textbf{0.00} & 1.69 & 0.7006 & 2.58 & 0.01 & 1.93 \\
& HRRR & \textbf{0.8961} & \textbf{1.57} & $-$0.17 & \textbf{1.08} & \textbf{0.8011} & \textbf{1.48} & $-$0.04 & \textbf{1.02} & \textbf{0.8966} & \textbf{1.53} & $-$0.08 & \textbf{1.09} \\
& GC   & 0.7950 & 2.30 & 0.71 & 1.64 & 0.6106 & 1.94 & 0.61 & 1.40 & 0.7418 & 2.43 & 0.81 & 1.76 \\
\midrule
\multirow{3}{*}{24h}
& EM   & 0.7930 & 2.70 & \textbf{0.00} & 2.08 & 0.6839 & 3.09 & \textbf{$-$0.01} & 2.34 & 0.7781 & 2.92 & \textbf{0.00} & 2.26 \\
& HRRR & \textbf{0.9426} & \textbf{1.46} & 0.10 & \textbf{1.07} & \textbf{0.9167} & \textbf{1.67} & 0.07 & \textbf{1.16} & \textbf{0.9430} & \textbf{1.50} & $-$0.06 & \textbf{1.09} \\
& GC   & 0.8462 & 2.52 & 0.76 & 1.78 & 0.7838 & 2.60 & 0.47 & 1.83 & 0.8441 & 2.62 & 0.14 & 1.93 \\
\midrule
\multirow{3}{*}{48h}
& EM   & 0.7673 & 3.38 & \textbf{0.00} & 2.54 & 0.7358 & 3.15 & \textbf{0.00} & 2.36 & 0.6014 & 3.19 & \textbf{$-$0.01} & 2.46 \\
& HRRR & \textbf{0.9426} & \textbf{1.70} & 0.21 & \textbf{1.17} & \textbf{0.9213} & \textbf{1.73} & $-$0.03 & \textbf{1.21} & \textbf{0.9092} & \textbf{1.59} & 0.30 & \textbf{1.12} \\
& GC   & 0.8360 & 2.89 & 0.58 & 1.86 & 0.8118 & 2.76 & 0.57 & 1.99 & 0.7406 & 2.47 & $-$0.24 & 1.82 \\
\bottomrule
\end{tabular}
\end{table*}

\begin{table*}[t]
\centering
\caption{10-m $v$-wind skill metrics (m\,s$^{-1}$) across three CONUS case studies. EM = AirCast-SR; GC = GraphCast. Bold $r$ indicates best correlation per lead time; bold Bias indicates smallest absolute bias.}
\label{tab:vgrd}
\begin{tabular}{llrrrrrrrrrrrr}
\toprule
& & \multicolumn{4}{c}{Dec 2022 (Winter)} & \multicolumn{4}{c}{Jun 2022 (Summer)} & \multicolumn{4}{c}{Mar 2023 (Spring)} \\
\cmidrule(lr){3-6} \cmidrule(lr){7-10} \cmidrule(lr){11-14}
Lead & Model & $r$ & RMSE & Bias & MAE & $r$ & RMSE & Bias & MAE & $r$ & RMSE & Bias & MAE \\
\midrule
\multirow{3}{*}{6h}
& EM   & 0.7469 & 2.71 & \textbf{0.00} & 2.04 & 0.3498 & 2.72 & \textbf{0.01} & 2.10 & 0.6760 & 2.90 & \textbf{0.00} & 2.23 \\
& HRRR & \textbf{0.9261} & \textbf{1.45} & $-$0.05 & \textbf{1.05} & \textbf{0.8223} & \textbf{1.45} & $-$0.12 & \textbf{1.03} & \textbf{0.9151} & \textbf{1.49} & $-$0.11 & \textbf{1.09} \\
& GC   & 0.5627 & 3.81 & 1.94 & 2.44 & 0.6438 & 1.95 & 0.67 & 1.43 & 0.8253 & 2.29 & 0.97 & 1.69 \\
\midrule
\multirow{3}{*}{24h}
& EM   & 0.6832 & 3.02 & \textbf{0.01} & 2.33 & 0.4662 & 3.34 & \textbf{0.01} & 2.56 & 0.6978 & 3.38 & \textbf{0.01} & 2.60 \\
& HRRR & \textbf{0.9311} & \textbf{1.39} & $-$0.06 & \textbf{1.03} & \textbf{0.8649} & \textbf{1.75} & $-$0.31 & \textbf{1.23} & \textbf{0.9372} & \textbf{1.56} & $-$0.21 & \textbf{1.16} \\
& GC   & 0.5700 & 4.09 & 2.42 & 2.93 & 0.7642 & 2.15 & 0.16 & 1.54 & 0.7198 & 3.50 & 1.49 & 2.27 \\
\midrule
\multirow{3}{*}{48h}
& EM   & 0.7230 & 2.75 & \textbf{0.00} & 2.03 & 0.5381 & 3.80 & \textbf{0.01} & 2.82 & 0.5230 & 3.60 & \textbf{$-$0.01} & 2.83 \\
& HRRR & \textbf{0.8958} & \textbf{1.78} & 0.21 & \textbf{1.24} & \textbf{0.9078} & \textbf{1.81} & $-$0.37 & \textbf{1.30} & \textbf{0.9100} & \textbf{1.54} & $-$0.18 & \textbf{1.12} \\
& GC   & 0.6437 & 3.43 & 1.80 & 2.40 & 0.8182 & 2.45 & 0.52 & 1.75 & 0.7860 & 2.60 & 0.79 & 1.91 \\
\bottomrule
\end{tabular}
\end{table*}

\begin{table}[b]
\centering
\caption{Zero-shot 2-m temperature evaluation ($r$ / RMSE [K]). EM = AirCast-SR; GC = GraphCast.}
\label{tab:global}
\begin{tabular}{lcccc}
\toprule
& \multicolumn{2}{c}{India (524 stn)} & \multicolumn{2}{c}{Germany (1338 stn)} \\
\cmidrule(lr){2-3} \cmidrule(lr){4-5}
Lead & EM & GC & EM & GC \\
\midrule
6h  & 0.88 / 3.74 & 0.95 / 2.85 & 0.63 / 1.54 & 0.78 / 1.14 \\
48h & 0.89 / 3.29 & 0.94 / 2.69 & 0.37 / 1.80 & 0.66 / 1.27 \\
\bottomrule
\end{tabular}
\end{table}

%% ============================================================
%% FIGURES
%% ============================================================

\begin{figure*}[t]
    \centering
    \includegraphics[width=\textwidth]{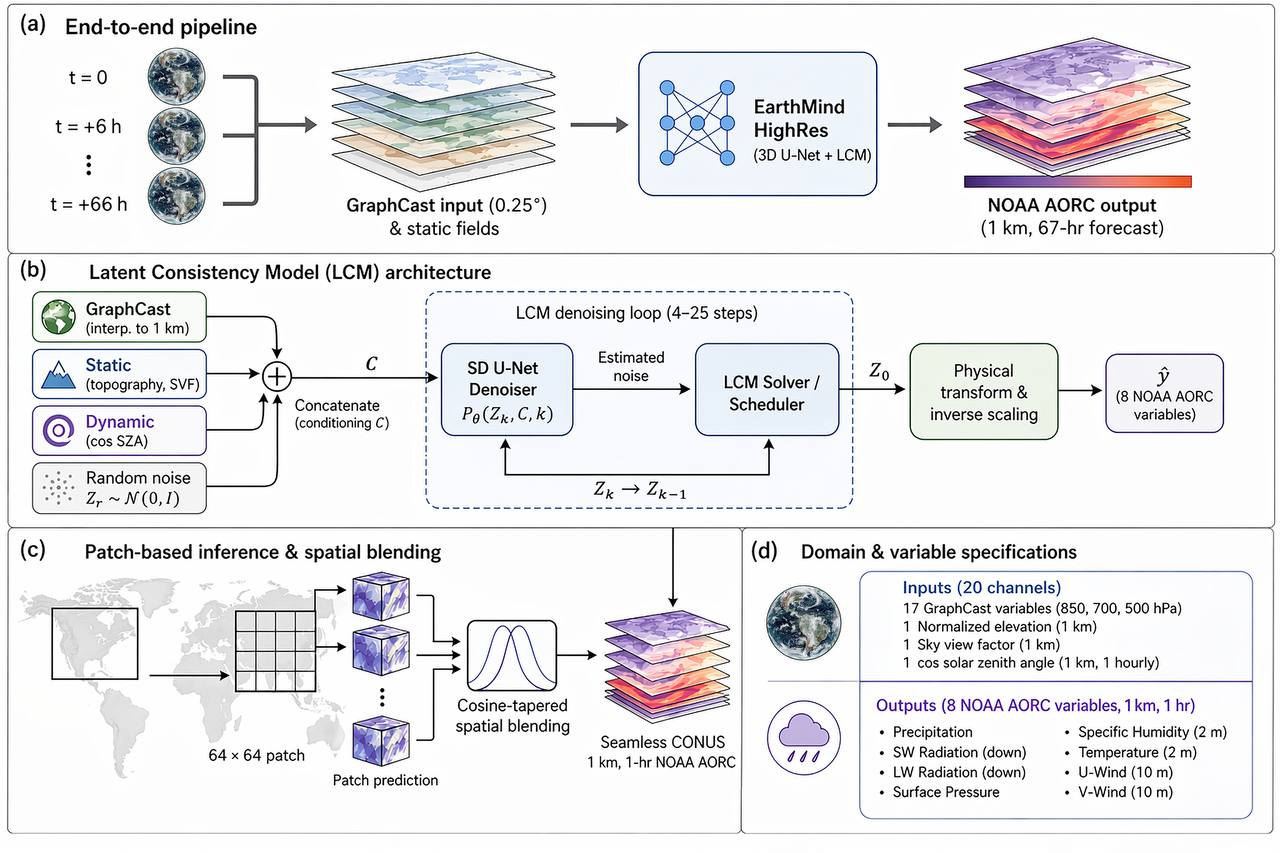}
    \caption{\textbf{AirCast-SR architecture and workflow.} (a) End-to-end pipeline. (b) Latent Consistency Model architecture. (c) Patch-based inference. (d) Input (20 channels) and output (7 NOAA AORC variables) specifications.}
    \label{fig:schematic}
\end{figure*}

\begin{thebibliography}{20}

\bibitem{lam2023graphcast}
R.~Lam, \emph{et al.},
``GraphCast: Learning skillful medium-range global weather forecasting,''
\emph{Science}, \textbf{382}, 1416--1421 (2023).

\bibitem{bi2023pangu}
K.~Bi, \emph{et al.},
``Accurate medium-range global weather forecasting with 3D neural networks,''
\emph{Nature}, \textbf{619}, 533--538 (2023).

\bibitem{chen2023fuxi}
L.~Chen, \emph{et al.},
``FuXi: A cascade machine learning forecasting system for 15-day global weather forecast,''
\emph{npj Climate and Atmospheric Science}, \textbf{6}, 190 (2023).

\bibitem{lang2024aifs}
S.~Lang, \emph{et al.},
``AIFS -- ECMWF's data-driven forecasting system,''
arXiv:2406.01465 (2024).

\bibitem{price2024gencast}
I.~Price, \emph{et al.},
``GenCast: Diffusion-based ensemble forecasting for medium-range weather,''
\emph{Nature}, \textbf{637}, 84--90 (2025).

\bibitem{bauer2015quiet}
P.~Bauer, A.~Thorpe, and G.~Brunet,
``The quiet revolution of numerical weather prediction,''
\emph{Nature}, \textbf{525}, 47--55 (2015).

\bibitem{dowell2022hrrr}
D.~C.~Dowell, \emph{et al.},
``The High-Resolution Rapid Refresh (HRRR): An hourly updating convection-allowing forecast model,''
\emph{Weather and Forecasting}, \textbf{37}, 1371--1395 (2022).

\bibitem{wood2004hydrologic}
A.~W.~Wood, \emph{et al.},
``Hydrologic implications of dynamical and statistical approaches to downscaling climate model outputs,''
\emph{Climatic Change}, \textbf{62}, 189--216 (2004).

\bibitem{hidalgo2008downscaling}
H.~G.~Hidalgo, \emph{et al.},
``Downscaling with constructed analogues: Daily precipitation and temperature fields over the United States,''
California Energy Commission Report CEC-500-2007-123 (2008).

\bibitem{vandal2017deepsd}
T.~Vandal, \emph{et al.},
``DeepSD: Generating high resolution climate change projections through single image super-resolution,''
in \emph{Proc.\ 23rd ACM SIGKDD International Conference on Knowledge Discovery and Data Mining (KDD)}, 1663--1672 (2017).

\bibitem{sha2020deep}
Y.~Sha, \emph{et al.},
``Deep-learning-based gridded downscaling of surface meteorological variables in complex terrain. Part I: Daily maximum and minimum 2-m temperature,''
\emph{Journal of Applied Meteorology and Climatology}, \textbf{59}(12), 2057--2073 (2020).

\bibitem{wang2021deep}
F.~Wang, \emph{et al.},
``Deep learning for daily precipitation and temperature downscaling,''
\emph{Water Resources Research}, \textbf{57}, e2020WR029308 (2021).

\bibitem{stengel2020adversarial}
K.~Stengel, \emph{et al.},
``Adversarial super-resolution of climatological wind and solar data,''
\emph{Proc.\ National Academy of Sciences}, \textbf{117}, 16805--16815 (2020).

\bibitem{harris2022generative}
L.~Harris, \emph{et al.},
``A generative deep learning approach to stochastic downscaling of precipitation forecasts,''
\emph{Journal of Advances in Modeling Earth Systems}, \textbf{14}, e2022MS003120 (2022).

\bibitem{mardani2025residual}
M.~Mardani, \emph{et al.},
``Residual corrective diffusion modeling for km-scale atmospheric downscaling,''
\emph{Communications Earth \& Environment}, \textbf{6}, 124 (2025);
preprint at arXiv:2309.15214 (2023).

\bibitem{ho2020denoising}
J.~Ho, \emph{et al.},
``Denoising diffusion probabilistic models,''
in \emph{Advances in Neural Information Processing Systems}, \textbf{33}, 6840--6851 (2020).

\bibitem{song2021scorebased}
Y.~Song, \emph{et al.},
``Score-based generative modeling through stochastic differential equations,''
in \emph{Proc.\ ICLR} (2021).

\bibitem{ronneberger2015unet}
O.~Ronneberger, \emph{et al.},
``U-Net: Convolutional networks for biomedical image segmentation,''
in \emph{Proc.\ MICCAI}, 234--241 (2015).

\bibitem{cciccek20163d}
\"{O}.~\c{C}i\c{c}ek, \emph{et al.},
``3D U-Net: Learning dense volumetric segmentation from sparse annotation,''
in \emph{Proc.\ MICCAI}, 424--432 (2016).

\bibitem{luo2023latent}
S.~Luo, \emph{et al.},
``Latent consistency models: Synthesizing high-resolution images with few-step inference,''
arXiv:2310.04378 (2023).

\bibitem{aorc2023}
NOAA Office of Water Prediction,
``Analysis of Record for Calibration (AORC), version 1.1,''
\url{https://hydrology.nws.noaa.gov/aorc-historic/} (2023).

\bibitem{pielawski2020coarse}
N.~Pielawski and C.~W\"{a}hlby,
``Introducing Hann windows for reducing edge-effects in patch-based image segmentation,''
\emph{PLoS ONE}, \textbf{15}, e0229839 (2020).

\bibitem{ecmwf2024stationbench}
ECMWF,
``StationBench: Surface station benchmark for AI weather models,''
software repository, \url{https://github.com/ecmwf-lab/stationbench} (accessed 2024).

\bibitem{schultz2021can}
M.~G.~Schultz, \emph{et al.},
``Can deep learning beat numerical weather prediction?,''
\emph{Phil.\ Trans.\ R.\ Soc.\ A}, \textbf{379}, 20200097 (2021).

\bibitem{blau2018perception}
Y.~Blau and T.~Michaeli,
``The perception--distortion tradeoff,''
in \emph{Proc.\ IEEE/CVF Conf.\ Computer Vision and Pattern Recognition (CVPR)}, 6228--6237 (2018).

\bibitem{song2023consistency}
Y.~Song, P.~Dhariwal, M.~Chen, and I.~Sutskever,
``Consistency models,''
in \emph{Proc.\ International Conference on Machine Learning (ICML)} (2023).

\end{thebibliography}
\end{document}